\documentclass{article}

\usepackage{times}

\usepackage{hyperref}
\usepackage{url}

\usepackage{amsmath}
\usepackage{amsfonts}
\usepackage{amsthm}
\usepackage{amssymb}
\usepackage{sansmath}

\usepackage{graphicx}
\graphicspath{ {images/} }
\usepackage{subcaption}
\usepackage{wrapfig}

\usepackage{natbib}
\usepackage{algorithm}
\usepackage{algorithmic}

\usepackage{booktabs}
\usepackage{relsize}
\usepackage{lipsum}
\usepackage{multirow}
\usepackage{rotating}
\usepackage{wrapfig}

\usepackage{inconsolata}
\usepackage[T1]{fontenc}
\usepackage{color}
\usepackage{verbatim}

\usepackage{hyperref}

\usepackage{soul}

\usepackage{enumitem}

\def\eg{{\it e.g,}}
\def\ben{\begin{equation}}
\def\een{\end{equation}}

\usepackage[accepted]{icml2017}

\icmltitlerunning{Gram-CTC}

\begin{document} 

\twocolumn[
\icmltitle{Gram-CTC: Automatic Unit Selection and Target Decomposition for Sequence Labelling}

\begin{icmlauthorlist}
\icmlauthor{Hairong Liu*}{ai}
\icmlauthor{Zhenyao Zhu*}{ai}
\icmlauthor{Xiangang Li}{ai}
\icmlauthor{Sanjeev Satheesh}{ai}
\end{icmlauthorlist}

\icmlaffiliation{ai}{Baidu Silicon Valley AI Lab,
            1195 Bordeaux Dr, Sunnyvale, CA 94089, USA}
            
\icmlcorrespondingauthor{Hairong Liu}{liuhairong@baidu.com}

\icmlkeywords{CTC, seq2seq}

\vskip 0.3in
]



\printAffiliationsAndNotice{\icmlEqualContribution} 

\begin{abstract}
Most existing sequence labelling models rely on a fixed decomposition of a target sequence into a sequence of basic units. These methods suffer from two major drawbacks: $1$) the set of basic units is fixed, such as the set of words, characters or phonemes in speech recognition, and $2$) the decomposition of target sequences is fixed. These drawbacks usually result in sub-optimal performance of modeling sequences. In this paper, we extend the popular CTC loss criterion to alleviate these limitations, and propose a new loss function called \textit{Gram-CTC}. While preserving the advantages of CTC, Gram-CTC automatically learns the best set of basic units (grams), as well as the most suitable decomposition of target sequences. 
Unlike CTC, Gram-CTC allows the model to output variable number of characters at each time step, which enables the model to capture longer term dependency and improves the computational efficiency.
We demonstrate that the proposed Gram-CTC improves CTC in terms of both performance and efficiency on the large vocabulary speech recognition task at multiple scales of data, and that with Gram-CTC we can outperform the state-of-the-art on a standard speech benchmark.

\end{abstract} 

\vspace{-10pt}
\section{Introduction}
\vspace{-4pt}
In recent years, there has been an explosion of interest in sequence labelling tasks. Connectionist Temporal Classification (CTC) loss \cite{graves2006connectionist} and Sequence-to-sequence (seq2seq) models \cite{cho2014learning, sutskever2014sequence} present powerful approaches to multiple applications, such as Automatic Speech Recognition (ASR) \cite{chan2016listen, Hannun2014DeepSS, bahdanau2016end}, machine translation \cite{cho2015using}, and parsing \cite{vinyals2015grammar}. These methods are based on $1)$ a fixed and carefully chosen set of basic units, such as words \cite{sutskever2014sequence}, phonemes \cite{chorowski2015attention} or characters \cite{chan2016listen}, and $2)$ a fixed and pre-determined decomposition of target sequences into these basic units. While these two preconditions greatly simplify the problems, especially the training processes, they are also strict and unnecessary constraints, which usually lead to suboptimal solutions. CTC models are especially harmed by fixed basic units in target space, because they build on the independence assumption between successive outputs in that space - an assumption which is often violated in practice.

The problem with fixed set of basic units is obvious: it is really hard, if not impossible, to determine the optimal set of basic units beforehand. For example, in English ASR, if we use words as basic units, we will need to deal with the large vocabulary-sized softmax, as well as rare words and data sparsity problem. On the other hand, if we use characters as basic units, the model is forced to learn the complex rules of English spelling and pronunciation. For example, the "oh" sound can be spelled in any of following ways, depending on the word it occurs in - \{ o, oa, oe, ow, ough, eau, oo, ew \}. While CTC can easily model commonly co-occuring grams together, it is impossible to give roughly equal probability to many possible spellings when transcribing unseen words. Most speech recognition systems model phonemes, sub-phoneme units and senones \eg ~ \cite{xiong2016microsoft} to get around these problems. Similarly, state-of-the-art neural machine translation systems use pre-segmented word pieces \eg ~\cite{Wu2016GooglesNM} aiming to find the best of both worlds. 

In reality, groups of characters are typically cohesive units for many tasks. For the ASR task, words can be decomposed into groups of characters that can be associated with sound (such as `tion' and `eaux'). For the machine translation task, there may be values in decomposing words as root words and extensions (so that meaning may be shared explicitly between `paternal' and `paternity'). Since this information is already available in the training data, it is perhaps, better to let the model figure it out by itself. At the same time, it raises another import question: how to decompose a target sequence into basic units? This is coupled with the problem of automatic selection of basic units, thus also better to let the model determine. Recently, there are some interesting attempts in these directions in the seq2seq framework. For example, Chan et al \cite{chan2016lsd} proposed the Latent Sequence Decomposition to decompose target sequences with variable length units as a function of both input sequence and the output sequence. 

In this work, we propose \textbf{Gram-CTC} - a strictly more general version of CTC - to automatically seek the best set of basic units from the training data, called \emph{grams}, and automatically decompose target sequences into sequences of grams. Just as sequence prediction with cross entropy training can be seen as special case of the CTC loss with a fixed alignment, CTC can be seen as a special case of Gram-CTC with a fixed decomposition of target sequences. Since it is a loss function, it can be applied to many seq2seq tasks to enable automatic selection of grams and decomposition of target sequences without modifying the underlying networks. 
Extensive experiments on multiple scales of data validate that Gram-CTC can improve CTC in terms of both performance and efficiency, and that using Gram-CTC the models outperform state-of-the-arts on standard speech benchmarks.



\vspace{-2pt}
\section{Related Work}
\vspace{-4pt}
The basic text units that previous works utilized for text prediction tasks (\eg, automatic speech recognition, handwriting recognition, machine translation, and image captioning) can be generally divided into two categories: hand-crafted ones and learning-based ones.

\textbf{Hand-crafted Basic Units.}
Fixed sets of characters (graphemes) \cite{graves2006connectionist, amodei2015deep}, word-pieces \cite{wu2016google, collobert2016wav2letter, zweig2016advances}, words \cite{soltau2016neural, cho2015using}, and phonemes \cite{lee1988large, sercu2016dense, xiong2016achieving} have been widely used as basic units for text prediction, but all of them have drawbacks. Using these fixed deterministic decompositions of text sequences defines a prior, which is not necessarily optimal for end-to-end learning.

\begin{itemize}[leftmargin=10pt]
\vspace{-10pt}
\item Word-segmented models remove the component of learning to spell and thus enable direct optimization towards reducing Word Error Rate (WER). However, these models suffer from having to handle a large vocabulary (1.7 million in \cite{soltau2016neural}), out-of-vocabulary words \cite{soltau2016neural, cho2015using} and data sparsity problems \cite{soltau2016neural}.  
\item Using characters results in much smaller vocabularies (\eg ~26 for English and thousands for Chinese), but it requires much longer contexts compared to using words or word-pieces and poses the challenge of composing characters to words \cite{graves2006connectionist, chan2015listen}, which is very noisy for languages like English.
\vspace{-5pt}
\item Word-pieces lie at the middle-ground of words and characters, providing a good trade-off between vocabulary size and context size, while the performance of using word pieces is sensitive to the choice of the word-piece set and its decomposition. 
\vspace{-5pt}
\item For the ASR task, the use of phonemes was popular in the past few decades as it eases acoustic modeling \cite{lee1988large} and good results were reported with phonemic models \cite{xiong2016achieving, sercu2016dense}. However, it introduces the uncertainties of mapping phonemes to words during decoding \cite{doss2003phoneme}, which becomes less robust especially for accented speech data.
\vspace{-5pt}
\end{itemize}

\textbf{Learning-based Basic Units.}
More recently, attempts have been made to learn basic unit sets automatically. \cite{luong2016achieving} proposed a  hybrid Word-Character model which translates mostly at the word level and consults the character components for rare words. Chan et al \cite{chan2016lsd} proposed the Latent Sequence Decompositions framework to decomposes target sequences with variable length-ed basic units as a function of both input sequence and the output sequence. 

There exist some earlier works on the ``unit discovery'' task \cite{cartwright1994segmenting,goldwater2006contextual}. A standard problem with MLE solutions to this task is that there are degenerate solutions, \textit{i.e.}, predicting the full corpus with probability $\mathbf{1}$ at the start. Often Bayesian priors or ``minimum description length'' constraints are used to remedy this.


\section{Gram-CTC}

\subsection{CTC}
CTC \cite{graves2006connectionist} is a very popular method in seq2seq learning since it does not require the alignment information between inputs and outputs, which is usually expensive, if not impossible, to obtain.

Since there is no alignment information, CTC marginalizes over all possible alignments. That is, it tries to maximize $p(l|x) = \sum_{\pi}p(\pi|x)$, where $x$ is input, and $\pi$ represent a valid alignment. For example, if the size of input is $3$, and the output is `hi', whose length is $2$, there are three possible alignments, `-hi', `h-i' and `hi-', where `-' represents \emph{blank}. For the details, please refer to the original paper \cite{graves2006connectionist}.

\vspace{-2pt}
\subsection{From CTC to Gram-CTC}
\vspace{-4pt}
\begin{figure*}[t]
\includegraphics[width=0.6\textwidth]{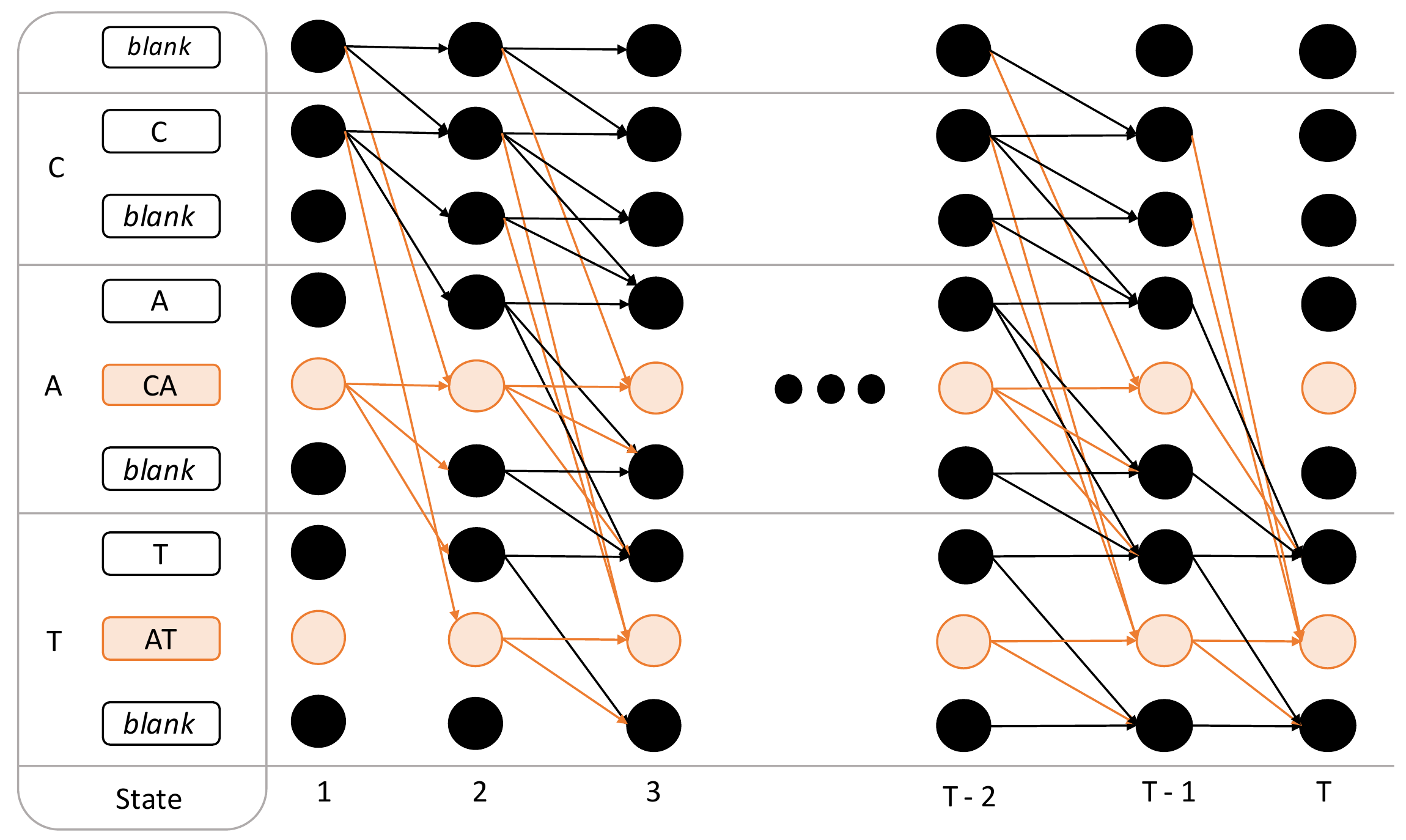}
\centering
\caption{Illustration of the states and the forward-backward transitions for the label `CAT'. Here we let $G$ be the set of all uni-grams and bi-grams of the English alphabet. The set of all valid states $S$ for the label $l$ = `CAT' are listed to the left. The set of states and transitions that are common to both vanilla and Gram-CTC are in black, and those that are unique to Gram-CTC are in orange. In general, any extension that collapses back to $l$ is a valid transition - For example, we can transition into (`CAT', 1) from  (`CAT', 1),  (`CA', 2),  (`CA', 1) and  (`CA', 0) but not from (`CAT', 0) or (`CAT', 2)}
\label{fig:dpgrams}
\end{figure*}
In CTC, the basic units are fixed, which is not desirable in some applications. Here we generalize CTC by considering a sequence of basic units, called \emph{gram}, as a whole, which is usually more reasonable in many applications.

Let $G$ be a set of n-grams of the set of basic units $C$ of the target sequence, and $\tau$ be the length of the longest gram in $G$. A Gram-CTC network has a softmax output layer with $|G|+1$ units, that is, the probability over all grams in $G$ and one additional symbol, \emph{blank}.
To simplify the problem, we also assume $C\subseteq G$. 
\footnote{This is because there may be no valid decompositions for some target sequences if $C\not \subseteq G$. Since Gram-CTC will figure out the ideal decomposition of target sequences into grams during training, this condition guarantees that there is at least one valid decomposition for every target sequence.}

For an input sequence $x$ of length $T$, let $y=N_{w}(x)$ be the sequence of network outputs, and denote by $y_k^t$ as the probability of the $k$-th gram at time $t$, where $k$ is the index of grams in $G^{\prime}=G\cup\{blank\}$,
then we have
\begin{equation}
p(\pi|x)=\prod_{t=1}^T y_{\pi_t}^t, \forall \pi\in {G^{\prime}}^T
\end{equation}

Just as in the case of CTC, here we refer to the elements of ${G^{\prime}}^T$ as paths, and denote them by $\pi$, which represents a possible alignment between input and output. The difference is that for each word in the target sequence, it may be decomposed into different sequences of grams. For example, the word `hello' can only be decomposed into the sequence [`h', `e', `l', `l', `o'] for CTC (assume uni-gram CTC here), but it also can be decomposed into the sequence [`he', `ll', `o'] if `he' and `ll' are in $G$.

For each $\pi$, we map it into a target sequence in the same way as CTC using the collapsing function that $1$) removes all repeated labels from the path and then $2$) removes all blanks. Note that essentially it is these rules which determine the transitions between the states of adjacent time steps in Figure \ref{fig:dpgrams}. This is a many-to-one mapping and we denote it by $B$. Note that other rules can be adopted here and the general idea presented in this paper does not depend on these specific rules. For a target sequence $l$, $B^{-1}(l)$ represents all paths mapped to $l$. Then, we have
\begin{equation}
p(l|x) = \sum_{\pi \in B^{-1}(l)}p(\pi|x)
\label{eqn:marginalization}
\end{equation}
This equation allows for training sequence labeling models without any alignment information using CTC loss, because it marginalizes over all possible alignments during training. Gram-CTC uses the same effect to enable the model to marginalize over not only alignments, but also decompositions of the target sequence.

Note that for each target sequence $l$, the set $B^{-1}(l)$ has $O(\tau^2)$ more paths than it does in CTC. This is because there are $O(\tau)$ times more valid states per time step, and each state may have a valid transition from $O(\tau)$ states in the previous time step. The original CTC method is thus, a special case of Gram-CTC when $G = C$ and $\tau = 1$. While the quadratic increase in the complexity of the algorithm is non trivial, we assert that it is a trivial increase in the overall training time of typical neural networks, where the computation time is dominated by the neural networks themselves. Additionally, the algorithm extends generally to any arbitrary $G$ and need not have all possible n-grams up to length $\tau$.

\subsection{The Forward-Backward Algorithm}

To efficiently compute $p(l|x)$, we also adopt the dynamic programming algorithm. The essence here is identifying the states of the problem, so that we may solve future states by reusing solutions to earlier states. In our case, the state must contain all the information required to identify all valid extensions of an incomplete path $\pi$ such that the collapsing function will eventually collapse the complete $\pi$ back to $l$. For Gram-CTC, this can be done by collapsing all but the last element of the path $\pi$. Therefore, the state is a tuple $(l_{1:i}, j)$, where the first item is a collapsed path, representing a prefix of the target label sequence, and $j \in \{0,\ldots,\tau\}$ is the length of the last gram $(l_{i-j+1:i})$ used for making the prefix. $j=0$ is valid and means that \emph{blank} was used. We denote the gram $(l_{i-j+1:i})$ by $g_i^j(l)$, and the state $(l_{1:i}, j)$ as $s_i^j(l)$. For readability, we will further shorten $s_i^j(l)$ to $s_i^j$ and $g_i^j(l)$ to $g_i^j$. For a state $s$, its corresponding gram is denoted by $s_g$, and the positions of the first character and last character of $s_g$ are denoted by $b(s)$ and $e(s)$, respectively. During dynamic programming, we are dealing with sequence of states, for a state sequence $\zeta$, its corresponding gram sequences is unique, denoted by $\zeta_g$.

Figure \ref{fig:dpgrams} illustrates partially the dynamic programming process for the target sequence `CAT'. Here we suppose $G$ contains all possible uni-grams and bi-grams. Thus, for each character in `CAT', there are three possible states associated with it: $1$) the current character, $2$) the bi-gram ending in current character, and $3)$ the \emph{blank} after current character. There is also one \emph{blank} at beginning. In total we have $10$ states. 



Supposing the maximum length of grams in $G$ is $\tau$, we first scan $l$ to get the set $S$ of all possible states, such that for all $s_i^j \in S$, its corresponding $g_i^j\in G^{\prime}$. $i\in\{0,\ldots,|l|\}$ and $j\in \{0,\ldots,\tau\}$. For a target sequence $l$, define the forward variable $\alpha_t(s)$ for any $s\in S$ to the total probability of all valid paths prefixes that end at state $s$ at time $t$.
\begin{equation}
\alpha_t(s)\stackrel{\text{def}}{=}\sum_{\zeta | B(\zeta_g)=l_{1:e(s)}, \zeta_t=s} \prod_{t'=1}^{t} y_{{\zeta_{t'}}_g}^{t'}
\end{equation}

Following this definition, we have the following rules for initialization
\begin{equation}
\alpha_1(s)=
\left \{
  \begin{array}{cl}
  y_b^1 & s = s_0^0 \\
  y_{g_i^i}^1 & s = s_i^i \quad \forall i\in\{1,\ldots,\tau\} \\
  0 & \text{otherwise}
  \end{array}
\right.
\end{equation}
and recursion
\begin{equation}
\alpha_t(s)=
\left \{
  \begin{array}{l}
  \hat{\alpha}_{t-1}^i*y^t_b \\ 
    ~~~~~~~~~~~~~~~~~~ \text{when } s = s_i^0,  \\
  \lbrack\hat{\alpha}_{t-1}^{i-j} + \alpha_{t-1}(s)\rbrack*y^t_{g_i^j}  \\ 
    ~~~~~~~~~~~~~~~~~~ \text{when }s = s_i^j\text{ and }g_i^j\neq g_{i-j}^j,  \\
  \lbrack\hat{\alpha}_{t-1}^{i-j} + \alpha_{t-1}(s) - \alpha_{t-1}(s_{i-j}^j)\rbrack*y^t_{g_i^j} \\ 
    ~~~~~~~~~~~~~~~~~~ \text{when }s = s_i^j\text{ and }g_i^j = g_{i-j}^j  
  \end{array}
\right.
\end{equation}
where $\hat{\alpha}_t^i = \sum_{j=0}^{\tau}\alpha_t(s_i^j)$ and $y^t_b$ is the probability of \emph{blank} at time $t$.

The total probability of the target sequence $l$ is then expressed in the following way:
\begin{equation}
p(l|x) = \sum_{j=0}^{\tau}\alpha_T(s_{|l|}^j)
\end{equation}

similarly, we can define the backward variable $\beta_t(s)$ as:
\begin{equation}
\beta_t(s)\stackrel{\text{def}}{=}\sum_{\zeta | B(\zeta_g)=l_{b(s):l}, \zeta_t=s} \prod_{t'=t}^{T} y_{{\zeta_{t'}}_g}^{t'}
\end{equation}

For the initialization and recursion of $\beta_t(s)$, we have
\begin{equation}
\beta_T(s)=
\left \{
  \begin{array}{cl}
  y_b^T & s = s_T^0 \\
  y_{g_T^i}^T & s = s_T^i \quad \forall i\in\{1,\ldots,\tau\} \\
  0 & \text{otherwise}
  \end{array}
\right.
\end{equation}
and
\begin{equation}
\beta_t(s)=
\left \{
  \begin{array}{ll}
  \hat{\beta}_{t+1}^i*y^t_b \\ 
    ~~~~~~~~~~~~~~~~~~ \text{when } s = s_i^0,  \\
  \lbrack\hat{\beta}_{t+1}^{i+j} + \beta_{t+1}(s)\rbrack*y^t_{g_i^j} \\ 
    ~~~~~~~~~~~~~~~~~~ \text{when }s = s_i^j\text{ and }g_i^j\neq g_{i+j}^j,  \\
  \lbrack\hat{\beta}_{t+1}^{i+j} + \beta_{t+1}(s) - \beta_{t+1}(s_{i+j}^j)\rbrack*y^t_{g_i^j} \\
    ~~~~~~~~~~~~~~~~~~ \text{when }s = s_i^j\text{ and }g_i^j = g_{i+j}^j  
  \end{array}
\right.
\end{equation}
where $\hat{\beta}_t^i = \sum_{j=0}^{\tau}\beta_t(s_{i+j}^j)$

\subsection{BackPropagation}
\vspace{-4pt}
Similar to CTC, we have the following expression:
\begin{equation}
p(l|x)=\sum_{s\in S}\frac{\alpha_t(s)\beta_t(s)}{y^t_{s_g}} \quad \forall t\in \{1,\ldots, T\}
\end{equation}

The derivative with regards to $y^t_k$ is:
\begin{equation}
\frac{\partial p(l|x)}{\partial y^t_k}=\frac{1}{{y^t_k}^2}\sum_{s\in lab(l,k)}\alpha_t(s)\beta_t(s)
\end{equation}
where $lab(l,k)$ is the set of states in $S$ whose corresponding gram is $k$. This is because there may be multiple states corresponding to the same gram.

For the backpropagation, the most important formula is the partial derivative of loss with regard to the unnormalized output $u^t_k$.

\begin{equation}
\frac{\partial \ln p(l|x)}{\partial u^t_k}=y^t_k-\frac{1}{y^t_k Z_t}\sum_{s\in lab(l,k)}\alpha_t(s)\beta_t(s)
\end{equation}
where $Z_t\stackrel{\text{def}}{=}\sum_{s\in S}\frac{\alpha_t(s)\beta_t(s)}{y^t_{s_g}}$.

\begin{figure*}[t]
\centering
\begin{subfigure}{.34\textwidth}
  \centering
  \includegraphics[width=1.0\linewidth]{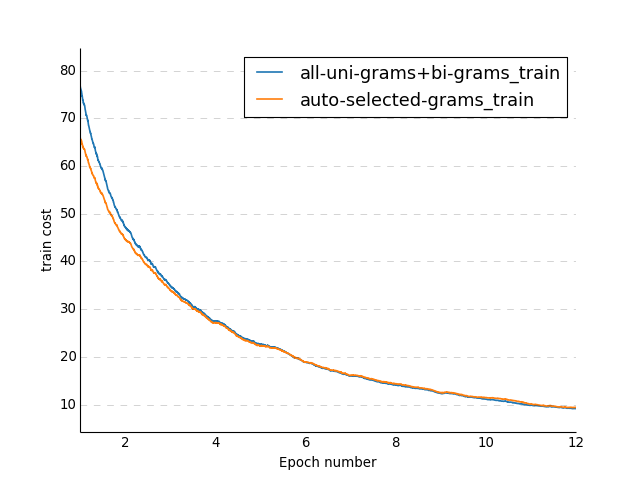}
  \caption{Training curves before ({\it blue}) and after ({\it orange}) auto-refinement of grams.}
  \label{fig:gram_selection}
\end{subfigure}
\begin{subfigure}{.34\textwidth}
  \centering
  \includegraphics[width=1.0\linewidth]{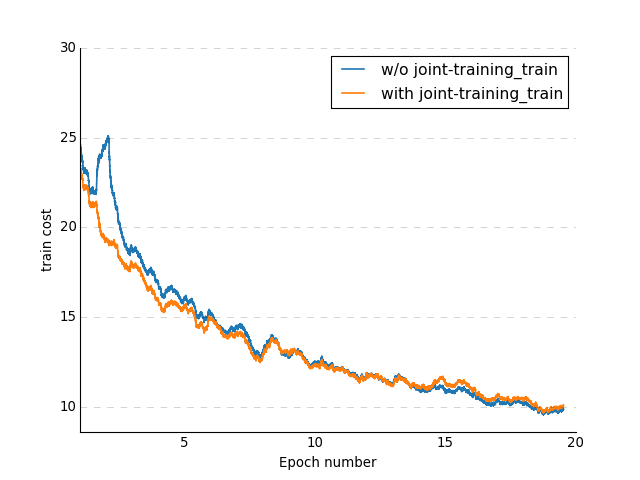}
  \caption{Training curves without ({\it blue}) and with ({\it orange}) joint-training}
  \label{fig:cotraining_curve}
\end{subfigure}
\begin{subfigure}{.28\textwidth}
  \centering
  \includegraphics[width=0.9\linewidth]{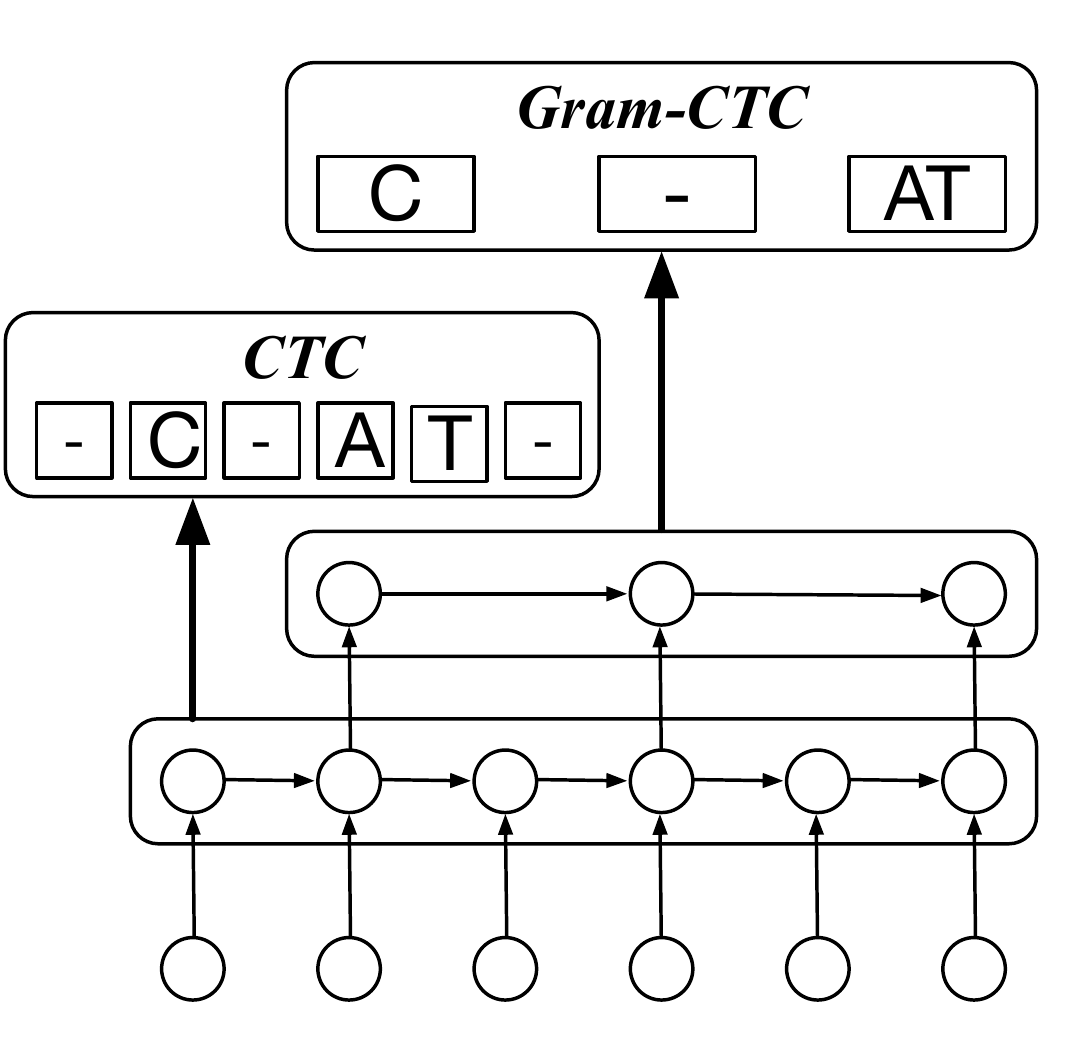}
  \caption{Joint-training Architecture}
  \label{fig:cotraining_model}
\end{subfigure}%
\caption{(Figure \ref{fig:gram_selection}) compares the training curves before (blue) and after (orange) auto-refinement of grams. They look very similar, although the number of grams is greatly reduced after refinement, which makes training faster and potentially more robust due to less gram sparsity. Figure (\ref{fig:cotraining_curve}) Training curve of model with and without joint-training. The model corresponding to the orange training curve is jointly trained together with vanilla CTC, such models are often more stable during training. Figure (\ref{fig:cotraining_model}) Typical joint-training model architecture - vanilla CTC loss is best applied a few levels lower than the Gram-CTC loss.\\
}
\vspace{-10pt}
\label{fig:fig}
\end{figure*}

\section{Methodology}

Here we describe additional techniques we found useful in practice to enable the Gram-CTC to work efficiently as well as effectively.

\subsection{Iterative Gram Selection}

Although Gram-CTC can automatically select useful grams, it is challenging to train with a large $G$. The total number of possible grams is usually huge. For example, in English, we have $26$ characters, then the total number of bi-grams is $26^2=676$, the total number of tri-grams are $26^3=17576$, \ldots, which grows exponentially and quickly becomes intractable. However, it is unnecessary to consider many grams, such as `aaaa', which are obviously useless.

In our experiments, we first eliminate most of useless grams from the statistics of a huge corpus, that is, we count the frequency of each gram in the corpus and drop these grams with rare frequencies. Then, we train a model with Gram-CTC on all the remaining grams. By applying (decoding) the trained model on a large speech dataset, we get the real statistics of gram's usage. Ultimately, we choose high frequency grams together with all uni-grams as our final gram set $G$.
Table~\ref{table:gram_selection} shows the impact of iterative gram selection on WSJ (without LM). Figure~\ref{fig:gram_selection} shows its corresponding training curve. For details, please refer to Section \ref{section:gram_selection}.

\subsection{Joint Training with Vanilla CTC}

Gram-CTC needs to solve both decomposition and alignment tasks, which is a harder task for a model to learn than CTC. This is often manifested in unstable training curves, forcing us to lower the learning rate which in turn results in models converging to a worse optima. To overcome this difficulty, we found it beneficial to train a model with both the Gram-CTC, as well as the vanilla CTC loss (similar to joint-training CTC together with CE loss as mentioned in \cite{Sak2015FastAA}). Joint training of multiple objectives for sequence labelling has also been explored in previous works \cite{kim2016joint, kim2016sequence}. 

A typical joint-training model looks like Figure~\ref{fig:cotraining_model}, and the corresponding training curve is shown in Figure~\ref{fig:cotraining_curve}.  The effect of joint-training are shown in Table \ref{table:fisher} and Table \ref{table:10k} in the experiments.

\section{Experiments}

We test the Gram-CTC loss on the ASR task, while both CTC and the introduced Gram-CTC are generic techniques for other sequence labelling tasks. For all of the experiments, the model specification and training procedure are the same as in \cite{amodei2015deep} - The model is a recurrent neural network (RNN) with 2 two-dimensional convolutional input layers, followed by K forward (Fwd) or bidirectional (Bidi) Gated Recurrent layers, N cells each, and one fully connected layer before a softmax layer. In short hand, such a model is written as `2x2D Conv - KxN GRU'. The network is trained end-to-end with the CTC, Gram-CTC or a weighted combination of both. This combination is described in the earlier section.

In all experiments, audio data is is sampled at 16kHz. Linear FFT features are extracted with a hop size of 10ms and window size of 20ms,  and are normalized so that each input feature has zero mean and unit variance. The network inputs are thus spectral magnitude maps ranging from 0-8kHz with 161 features per 10ms frame.  At each epoch, $40\%$ of the utterances are randomly selected to add background noise to. The optimization method we use is stochastic gradient descent with Nesterov momentum. Learning hyperparameters (batch-size, learning-rate, momentum, and etc.) vary across different datasets and are tuned for each model by optimizing a hold-out set. Typical values are a learning rate of $10^{-3}$ and momentum of $0.99$.

\subsection{Data and Setup}

{\bf Wall Street Journal (WSJ)}. This corpora consists primarily of read speech with texts drawn from a machine-readable corpus of Wall Street Journal news text, and contains about {\it 80 hours} speech data. We used the standard configuration of train si284 dataset for training, dev93 for validation and eval92 for testing. This is a relatively `clean' task and often used for model prototyping \cite{miao2015eesen, bahdanau2016end, zhang2016very, chan2016lsd}.

{\bf Fisher-Switchboard}.  This is a commonly used English conversational telephone speech (CTS) corpora, which contains {\it 2300 hours} CTS data. Following the previous works \cite{geoffery2016ctc,povey2016purely,xiong2016achieving,sercu2016dense}, evaluation is carried out on the NIST 2000 CTS test set, which comprises both Switchboard (SWB) and CallHome (CH) subsets.

{\bf 10K Speech Dataset}. We conduct large scale ASR experiments on a noisy internal dataset of {\it 10,000 hours}. This dataset contains speech collected from various scenarios, such as different background noises, far-field, different accents, and so on. Due to its inherent complexities, it is a very challenging task, and can thus validate the effectiveness of the proposed method for real-world application.

\begin{table}[t]
\begin{center}
\begin{tabular}{l|c}
\toprule
Loss & WER \\
\midrule
CTC, uni-gram & 16.91 \\
CTC, bi-gram & 21.63 \\
\midrule
Gram-CTC, handpick & 17.01 \\
Gram-CTC, all uni-grams + bi-grams & 16.89 \\
Gram-CTC, auto-refinement & \textbf{16.66} \\
\bottomrule
\end{tabular}
\end{center}
\vspace{-10pt}
\caption{Results of different gram selection methods on WSJ dataset.}
\label{table:gram_selection}
\vspace{-10pt}
\end{table}

\subsection{Gram Selection}

\label{section:gram_selection}
We employ the WSJ dataset for demonstrating different strategies of selecting grams for Gram-CTC, since it is a widely used dataset and also small enough for rapid idea verification. However, because it is small, we cannot use large grams here due to data sparsity problem. Thus, the auto-refined gram set on WSJ is not optimal for other larger datasets, where larger grams could be effectively used, but the procedure of refinement is the same for them.

We first train a model using all uni-grams and bi-grams ($29$ uni-grams and $26^2=676$ bi-grams, in total $705$ grams), and then do decoding with the obtained model on another speech dataset to get the statistics of the usage of grams. Top $100$ bi-grams together with all $29$ uni-grams (auto-refined grams) are used for the second round of training. For comparison, we also present the result of the best hand-picked grams, as well as the results on uni-grams. All the results are shown in Table \ref{table:gram_selection}.

Some interesting observations can be found in Table \ref{table:gram_selection}. First, the performance of auto-refined grams is only slightly better than the combination of all uni-grams and all bi-grams. This is probably because WSJ is so small that gram learning suffers from the data sparsity problem here (similar to word-segmented models). The auto-refined gram set contains only a small subset of bi-grams, thus more robust. This is also why we only try bi-grams, not including higher-order grams. Second, the performance of best handpicked grams is worse than auto-refined grams. This is desirable. It is time-consuming to handpick grams, especially when you consider high-order grams. The method of iterative gram selection is not only fast, but usually better. Third, the performance of Gram-CTC on auto-refined grams is only slightly better than CTC on uni-grams. 
This is because Gram-CTC is inherently difficult to train, since it needs to learn both decomposition and alignment. WSJ is too small to provide enough data to train Gram-CTC.

\begin{table}[t]
\begin{center}
\begin{tabular}{l| c c | c  c}
\toprule
Loss & \multicolumn{2}{c|}{WER} & \multicolumn{2}{c}{\small{Epoch Time (mins)}}  \\
\multicolumn{1}{r|}{(stride)}    &  2 &  4 &  2 &  4 \\
\midrule
CTC, uni-gram &  16.91 & 23.76 & ~~29~ & ~~16~\\
CTC, bi-gram &  20.57 & 21.63 & ~~23~ & ~~12~ \\
Gram-CTC &  16.66 &  18.87 & ~~35~ & ~~18~\\
\bottomrule
\end{tabular}
\end{center}
\vspace{-10pt}
\caption{Performances with different model strides on WSJ dataset.}
\label{table:stride}
\vspace{-10pt}
\end{table}
\vspace{-8pt}

\begin{figure*}[ht]
\centering
\fbox{
\includegraphics[width=0.97\textwidth]{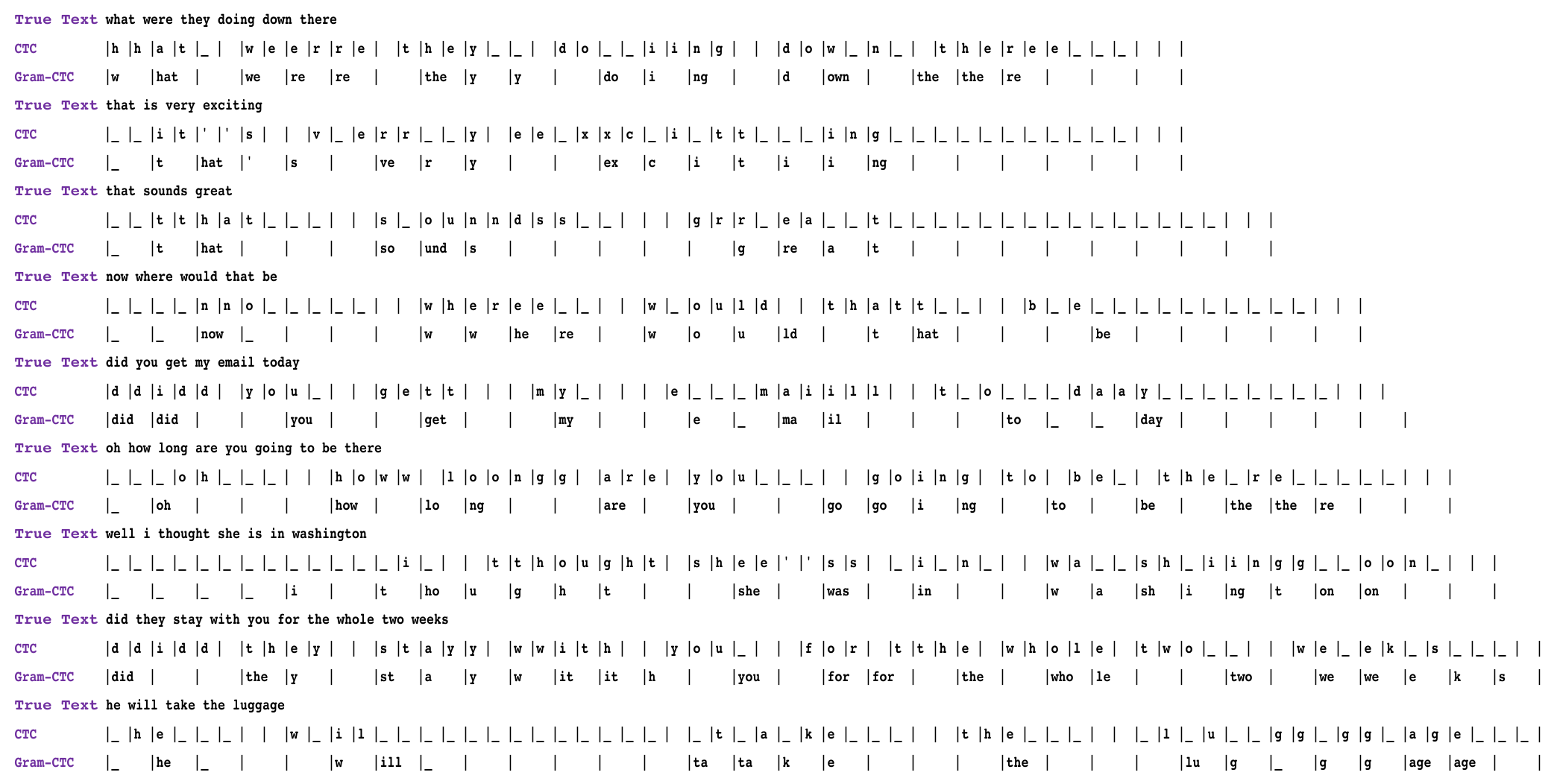}
}
\caption{Max-decoding results (without collapsing) of CTC and Gram-CTC on utterances from Switchboard dataset. The predicted characters (by CTC) or grams  (by Gram-CTC) at each timestep are separated by "|". As the Gram-CTC model is trained with doubled stride as that of CTC model, we place the grams at a doubled width as we do with characters for better viewing. The "\_" represents {\it blank}. }
\label{fig:examples}
\end{figure*}

\subsection{Sequence Labelling in Large Stride}

Using a large time stride for sequence labelling with RNNs can greatly boost the overall computation efficiency, since it effectively reduces the time steps for recurrent computation, thus speeds up the process of both forward inference and backward propagation. However, the largest stride that can be used is limited by the gram set we use. The (uni-gram) CTC has to work in a high time resolution (small stride) in order to have enough number of frames to output every character. This is very inefficient as we know the same acoustic feature could correspond to several grams of different lengths (\textit{e.g.}, \{`i', `igh', `eye'\}) . The larger the grams are, the larger stride we are potentially able to use.

DS2 \cite{amodei2015deep} employed non-overlapping bi-gram outputs to allow for a larger stride. This imposes an artificial constraint forcing the model to learn, not only the spelling of each word, but also how to split words into bi-grams. For example, \textit{part} is split as [\textit{pa}, \textit{rt}] but the word \textit{apart} is forced to be decomposed as [\textit{ap}, \textit{ar}, \textit{t}]. Gram-CTC removes this constraint by allowing the model to decompose words into larger units into the most convenient or sensible decomposition. Comparison results show this change enables Gram-CTC to work much better than bi-gram CTC, as in Table \ref{table:stride}.

In Table \ref{table:stride}, we compare the performance of trained model and training efficiency on two strides, $2$ and $4$. For Gram-CTC, we use the auto-refined gram set from previous section. As expected, using stride $4$ almost cuts the training time per epoch into half, compared to stride $2$. 
From stride $2$ to stride $4$, the performance of uni-gram CTC drops quickly. This is because small grams inherently need higher time resolutions. 
As for Gram-CTC, from stride $2$ to stride $4$, its performance decreases a little bit, while in experiments on the other datasets, Gram-CTC constantly works better in stride $4$. 
One possible explanation is that WSJ is too small for Gram-CTC to learn large grams well.
In contrast, the performance of bi-gram CTC is not as good as that of Gram-CTC in either stride.

\subsection{Decoding Examples}

Figure \ref{fig:examples} illustrates the max-decoding results of both CTC and Gram-CTC on nine utterances. Here the label set for CTC is the set of all characters, and the label set for Gram-CTC is an auto-refined gram set containing all uni-grams and some high-frequency high-order grams. Here Gram-CTC uses stride $4$ while CTC uses stride $2$.

From Figure \ref{fig:examples}, we can find that: $1)$ Gram-CTC does automatically find many intuitive and meaningful grams, such as `the', `ng', and `are'. $2)$ It also decomposes the sentences into segments which are meaningful in term of pronunciation. This decomposition resembles the phonetic decomposition, but in larger granularity and arguably more natural. $3)$ Since Gram-CTC predicts a chunk of characters (a gram) each time, each prediction utilizes larger context and these characters in the same predicted chunk are dependent, thus potentially more robust. One example is the word `will' in the last sentence in Figure \ref{fig:examples}. $4)$ Since the output of network is the probability over all grams, the decoding process is almost the same as CTC, still end-to-end. This makes such decomposition superior to phonetic decomposition. In summary, Gram-CTC combines the advantages of both CTC on characters and CTC on phonemes.

\begin{table}[t]
\begin{center}
\small{
\begin{tabular}{l|c}
\toprule
Architecture &  WER  \\
\midrule
Phoneme CTC + trigram LM  \cite{miao2015eesen}  & 7.3 \\
Grapheme CTC + trigram LM  \cite{miao2015eesen}  & 9.0 \\
Attention + trigram LM   \cite{bahdanau2016end} &   9.3 \\ 
DeepConv LAS + no LM  \cite{zhang2016very} & 10.5 \\
DeepConv LAS + LSD + no LM  \cite{chan2016lsd}  & \textbf{9.6} \\
Temporal LS + Cov + LM \\~~\cite{chorowski2016seq2seq} & \textbf{6.7} \\
\midrule
Vanilla CTC + no LM  (ours)  & 16.91  \\
Vanilla CTC + LM  (ours)  & 7.11 \\ 
\midrule
Gram-CTC + no LM (ours)  & 16.66 \\
Gram-CTC + LM (ours)  & \textbf{6.75} \\
\bottomrule
\end{tabular}
}
\end{center}
\vspace{-10pt}
\caption{Comparison with previous published results with end-to-end training on WSJ speech dataset. The numbers in bold are the best results with and without a language model}
\label{table:wsj_sota}
\end{table}

\subsection{Comparison with Other Methods}

\subsubsection{WSJ dataset}

The model used here is [2x2D conv, 3x1280 Bidi GRU] with a CTC or Gram-CTC loss. The results are shown in Table \ref{table:wsj_sota}. For all models we trained, language model can greatly improve their performances, in term of WER. Though this dataset contains very limited amount of text data for learning gram selection and decomposition, Gram-CTC can still improve the vanilla CTC notably.

\subsubsection{Fisher-Switchboard}

The acoustic model trained here is composed of two 2D convolutions and six bi-directional GRU layer in 2048 dimension. The corresponding labels are used for training N-gram language models.
\begin{itemize}
\vspace{-10pt}
\item Switchboard English speech 97S62
\vspace{-7pt}
\item Fisher English speech Part 1 - 2004S13, 2004T19
\vspace{-7pt}
\item Fisher English speech Part 2 - 2005S13, 2005T19
\vspace{-10pt}
\end{itemize}
We use a sample of the Switchboard-1 portion of the NIST 2002 dataset (2004S11 RT-02) for tuning language model hyper-parameters. The evaluation is done on the NIST 2000 set. This configuration forms a standard benchmark for evaluating ASR models. Results are in Table~\ref{table:fisher}.

We compare our model against best published results on {\it in-domain} data. These results can often be improved using {\it out-of-domain} data for training the language model, and sometimes the acoustic model as well. Together these techniques allow \cite{xiong2016achieving} to reach a WER of 5.9 on the SWBD set.


\subsubsection{10K Speech Dataset}

Finally, we experiment on a large noisy dataset collected by ourself for building large-vocabulary Continuous Speech Recognition (LVCSR) systems. This dataset contains about $10000$ hours speech in a diversity of scenarios, such as far-field, background noises, accents.  In all cases, the model is [2x2D Conv, 3x2560 Fwd GRU, LA Conv] with only a change in the loss function. `LA Conv' refers to a look ahead convolution layer as seen in \cite{amodei2015deep} which works together with forward-only RNNs for deployment purpose. 

As with the Fisher-Switchboard dataset, the optimal stride is $4$ for Gram-CTC and $2$ for vanilla CTC on this dataset. Thus, in both experiments, both Gram-CTC and vanilla CTC + Gram-CTC are trained mush faster than vanilla CTC itself. The result is shown in Table \ref{table:10k}. Gram-CTC performs better than CTC. After joint-training with vanilla CTC and alignment information through a CE loss, its performance is further boosted, which verifies joint-training helps training. In fact, with only a small additional cost of time, it effectively reduces the WER from $27.56\%$ to $25.59\%$ (without language model).


\begin{table}[t]
\begin{center}
\small{
\begin{tabular}{l|c|c}
\toprule
Architecture &  SWBD & CH  \\
  & WER & WER  \\
\midrule
Iterated-CTC  \cite{geoffery2016ctc} & 11.3 & 18.7 \\ 
BLSTM + LF MMI  \cite{povey2016purely} & 8.5 & 15.3 \\ 
LACE + LF MMI \footnote{An unreported result using RNN-LM trained on in-domain text could be better than this result}
 \cite{xiong2016achieving} & 8.3 & 14.8 \\ 
Dilated convolutions  \cite{sercu2016dense} & 7.7 & \textbf{14.5} \\ 
\midrule
Vanilla CTC  (ours) & 9.0 & 17.7   \\ 
Gram-CTC  (ours) & 7.9 & 15.8  \\ 
Vanilla CTC + Gram-CTC  (ours) & \textbf{7.3} & 14.7  \\ 
\bottomrule
\end{tabular}
}
\end{center}
\vspace{-10pt}
\caption{Comparison with previous published results on Fisher-Switchboard benchmark (``SWBD'' and ``CH'' represent Switchboard and Callhome portions, respectively) using {\it in-domain} data. We only list results using single models here.}
\label{table:fisher}
\end{table}

\begin{table}[t]
\small{
\begin{tabular}{l|c|c}
\toprule
Architecture & WER(No LM) & WER(With LM) \\
\midrule
Vanilla CTC  & 29.1 & 19.77 \\ 
Gram-CTC & 27.56 & 19.53 \\ 
Vanilla CTC + Gram-CTC & 25.59 & \textbf{18.52} \\ 
\bottomrule
\end{tabular}
}
\vspace{-5pt}
\caption{LVCSR results on 10K speech dataset.}
\label{table:10k}
\end{table}

\section{Conclusions and Future Work}

In this paper, we have proposed the \textit{Gram-CTC} loss to enable automatic decomposition of target sequences into learned grams. 
We also present techniques to train the Gram-CTC in a clean and stable way. 
Our extensive experiments demonstrate the proposed Gram-CTC enables the models to run more efficiently than the vanilla CTC, by using larger stride, while obtaining better performance of sequence labelling.
Comparison experiments on multiple-scale datasets show the proposed Gram-CTC obtains state-of-the-art results on various ASR tasks. 

An interesting observation is that the learning of Gram-CTC implicitly avoids the ``degenerated solution'' that occurring in the traditional ``unit discovery'' task, without involving any Bayesian priors or the ``minimum description length'' constraint. Using a small gram set that contains only short (up to $5$ in our experiments) as well as high-frequency grams may explain the success here. 

We will continue investigating techniques of improving the optimization of Gram-CTC loss, as well as the applications of Gram-CTC for other sequence labelling tasks.

\clearpage
\bibliographystyle{icml2017}
\bibliography{main}

\begin{thebibliography}{33}
\providecommand{\natexlab}[1]{#1}
\providecommand{\url}[1]{\texttt{#1}}
\expandafter\ifx\csname urlstyle\endcsname\relax
  \providecommand{\doi}[1]{doi: #1}\else
  \providecommand{\doi}{doi: \begingroup \urlstyle{rm}\Url}\fi

\bibitem[Graves et~al.(2006)Graves, Fern{\'a}ndez, Gomez, and
  Schmidhuber]{graves2006connectionist}
Alex Graves, Santiago Fern{\'a}ndez, Faustino Gomez, and J{\"u}rgen
  Schmidhuber.
\newblock Connectionist temporal classification: labelling unsegmented sequence
  data with recurrent neural networks.
\newblock In \emph{Proceedings of the 23rd international conference on Machine
  learning}, pages 369--376. ACM, 2006.

\bibitem[Cho et~al.(2014)Cho, Van~Merri{\"e}nboer, Gulcehre, Bahdanau,
  Bougares, Schwenk, and Bengio]{cho2014learning}
Kyunghyun Cho, Bart Van~Merri{\"e}nboer, Caglar Gulcehre, Dzmitry Bahdanau,
  Fethi Bougares, Holger Schwenk, and Yoshua Bengio.
\newblock Learning phrase representations using rnn encoder-decoder for
  statistical machine translation.
\newblock \emph{arXiv preprint arXiv:1406.1078}, 2014.

\bibitem[Sutskever et~al.(2014)Sutskever, Vinyals, and
  Le]{sutskever2014sequence}
Ilya Sutskever, Oriol Vinyals, and Quoc~V Le.
\newblock Sequence to sequence learning with neural networks.
\newblock In \emph{Advances in neural information processing systems}, pages
  3104--3112, 2014.

\bibitem[Chan et~al.(2016{\natexlab{a}})Chan, Jaitly, Le, and
  Vinyals]{chan2016listen}
William Chan, Navdeep Jaitly, Quoc Le, and Oriol Vinyals.
\newblock Listen, attend and spell: A neural network for large vocabulary
  conversational speech recognition.
\newblock In \emph{2016 IEEE International Conference on Acoustics, Speech and
  Signal Processing (ICASSP)}, pages 4960--4964. IEEE, 2016{\natexlab{a}}.

\bibitem[Hannun et~al.(2014)Hannun, Case, Casper, Catanzaro, Diamos, Elsen,
  Prenger, Satheesh, Sengupta, Coates, and Ng]{Hannun2014DeepSS}
Awni~Y. Hannun, Carl Case, Jared Casper, Bryan Catanzaro, Greg Diamos, Erich
  Elsen, Ryan Prenger, Sanjeev Satheesh, Shubho Sengupta, Adam Coates, and
  Andrew~Y. Ng.
\newblock Deep speech: Scaling up end-to-end speech recognition.
\newblock \emph{CoRR}, abs/1412.5567, 2014.

\bibitem[Bahdanau et~al.(2016)Bahdanau, Chorowski, Serdyuk, Bengio,
  et~al.]{bahdanau2016end}
Dzmitry Bahdanau, Jan Chorowski, Dmitriy Serdyuk, Yoshua Bengio, et~al.
\newblock End-to-end attention-based large vocabulary speech recognition.
\newblock In \emph{2016 IEEE International Conference on Acoustics, Speech and
  Signal Processing (ICASSP)}, pages 4945--4949. IEEE, 2016.

\bibitem[S{\'e}bastien et~al.(2015)S{\'e}bastien, Cho, Memisevic, and
  Bengio]{cho2015using}
Jean S{\'e}bastien, Kyunghyun Cho, Roland Memisevic, and Yoshua Bengio.
\newblock On using very large target vocabulary for neural machine translation.
\newblock 2015.

\bibitem[Vinyals et~al.(2015)Vinyals, Kaiser, Koo, Petrov, Sutskever, and
  Hinton]{vinyals2015grammar}
Oriol Vinyals, {\L}ukasz Kaiser, Terry Koo, Slav Petrov, Ilya Sutskever, and
  Geoffrey Hinton.
\newblock Grammar as a foreign language.
\newblock In \emph{Advances in Neural Information Processing Systems}, pages
  2773--2781, 2015.

\bibitem[Chorowski et~al.(2015)Chorowski, Bahdanau, Serdyuk, Cho, and
  Bengio]{chorowski2015attention}
Jan~K Chorowski, Dzmitry Bahdanau, Dmitriy Serdyuk, Kyunghyun Cho, and Yoshua
  Bengio.
\newblock Attention-based models for speech recognition.
\newblock In \emph{Advances in Neural Information Processing Systems}, pages
  577--585, 2015.

\bibitem[Xiong et~al.(2016{\natexlab{a}})Xiong, Droppo, Huang, Seide, Seltzer,
  Stolcke, Yu, and Zweig]{xiong2016microsoft}
W~Xiong, J~Droppo, X~Huang, F~Seide, M~Seltzer, A~Stolcke, D~Yu, and G~Zweig.
\newblock The microsoft 2016 conversational speech recognition system.
\newblock \emph{arXiv preprint arXiv:1609.03528}, 2016{\natexlab{a}}.

\bibitem[Wu et~al.(2016{\natexlab{a}})Wu, Schuster, Chen, Le, Norouzi,
  Macherey, Krikun, Cao, Gao, Macherey, Klingner, Shah, Johnson, Liu, Kaiser,
  Gouws, Kato, Kudo, Kazawa, Stevens, Kurian, Patil, Wang, Young, Smith, Riesa,
  Rudnick, Vinyals, Corrado, Hughes, and Dean]{Wu2016GooglesNM}
Yonghui Wu, Mike Schuster, Zhifeng Chen, Quoc~V. Le, Mohammad Norouzi, Wolfgang
  Macherey, Maxim Krikun, Yuan Cao, Qin Gao, Klaus Macherey, Jeff Klingner,
  Apurva Shah, Melvin Johnson, Xiaobing Liu, Lukasz Kaiser, Stephan Gouws,
  Yoshikiyo Kato, Taku Kudo, Hideto Kazawa, Keith Stevens, George Kurian,
  Nishant Patil, Wei Wang, Cliff Young, Jason Smith, Jason Riesa, Alex Rudnick,
  Oriol Vinyals, Gregory~S. Corrado, Macduff Hughes, and Jeffrey Dean.
\newblock Google's neural machine translation system: Bridging the gap between
  human and machine translation.
\newblock \emph{CoRR}, abs/1609.08144, 2016{\natexlab{a}}.

\bibitem[Chan et~al.(2016{\natexlab{b}})Chan, Zhang, Le, and
  Jaitly]{chan2016lsd}
William Chan, Yu~Zhang, Quoc Le, and Navdeep Jaitly.
\newblock Latent sequence decompositions.
\newblock In \emph{Arxiv}, 2016{\natexlab{b}}.

\bibitem[Amodei et~al.(2015)Amodei, Anubhai, Battenberg, Case, Casper,
  Catanzaro, Chen, Chrzanowski, Coates, Diamos, et~al.]{amodei2015deep}
Dario Amodei, Rishita Anubhai, Eric Battenberg, Carl Case, Jared Casper, Bryan
  Catanzaro, Jingdong Chen, Mike Chrzanowski, Adam Coates, Greg Diamos, et~al.
\newblock Deep speech 2: End-to-end speech recognition in english and mandarin.
\newblock \emph{arXiv preprint arXiv:1512.02595}, 2015.

\bibitem[Wu et~al.(2016{\natexlab{b}})Wu, Schuster, Chen, Le, Norouzi,
  Macherey, Krikun, Cao, Gao, Macherey, et~al.]{wu2016google}
Yonghui Wu, Mike Schuster, Zhifeng Chen, Quoc~V Le, Mohammad Norouzi, Wolfgang
  Macherey, Maxim Krikun, Yuan Cao, Qin Gao, Klaus Macherey, et~al.
\newblock Google's neural machine translation system: Bridging the gap between
  human and machine translation.
\newblock \emph{arXiv preprint arXiv:1609.08144}, 2016{\natexlab{b}}.

\bibitem[Collobert et~al.(2016)Collobert, Puhrsch, and
  Synnaeve]{collobert2016wav2letter}
Ronan Collobert, Christian Puhrsch, and Gabriel Synnaeve.
\newblock Wav2letter: an end-to-end convnet-based speech recognition system.
\newblock \emph{arXiv preprint arXiv:1609.03193}, 2016.

\bibitem[Zweig et~al.(2016{\natexlab{a}})Zweig, Yu, Droppo, and
  Stolcke]{zweig2016advances}
Geoffrey Zweig, Chengzhu Yu, Jasha Droppo, and Andreas Stolcke.
\newblock Advances in all-neural speech recognition.
\newblock \emph{arXiv preprint arXiv:1609.05935}, 2016{\natexlab{a}}.

\bibitem[Soltau et~al.(2016)Soltau, Liao, and Sak]{soltau2016neural}
Hagen Soltau, Hank Liao, and Hasim Sak.
\newblock Neural speech recognizer: Acoustic-to-word lstm model for large
  vocabulary speech recognition.
\newblock \emph{arXiv preprint arXiv:1610.09975}, 2016.

\bibitem[Lee and Hon(1988)]{lee1988large}
K-F Lee and H-W Hon.
\newblock Large-vocabulary speaker-independent continuous speech recognition
  using hmm.
\newblock In \emph{Acoustics, Speech, and Signal Processing, 1988. ICASSP-88.,
  1988 International Conference on}, pages 123--126. IEEE, 1988.

\bibitem[Sercu and Goel(2016)]{sercu2016dense}
Tom Sercu and Vaibhava Goel.
\newblock Dense prediction on sequences with time-dilated convolutions for
  speech recognition.
\newblock \emph{arXiv preprint arXiv:1611.09288}, 2016.

\bibitem[Xiong et~al.(2016{\natexlab{b}})Xiong, Droppo, Huang, Seide, Seltzer,
  Stolcke, Yu, and Zweig]{xiong2016achieving}
Wayne Xiong, Jasha Droppo, Xuedong Huang, Frank Seide, Mike Seltzer, Andreas
  Stolcke, Dong Yu, and Geoffrey Zweig.
\newblock Achieving human parity in conversational speech recognition.
\newblock \emph{arXiv preprint arXiv:1610.05256}, 2016{\natexlab{b}}.

\bibitem[Chan et~al.(2015)Chan, Jaitly, Le, and Vinyals]{chan2015listen}
William Chan, Navdeep Jaitly, Quoc~V Le, and Oriol Vinyals.
\newblock Listen, attend and spell.
\newblock \emph{arXiv preprint arXiv:1508.01211}, 2015.

\bibitem[Doss et~al.(2003)Doss, Stephenson, Bourlard, and
  Bengio]{doss2003phoneme}
Mathew~Magimai Doss, Todd~A Stephenson, Herv{\'e} Bourlard, and Samy Bengio.
\newblock Phoneme-grapheme based speech recognition system.
\newblock In \emph{Automatic Speech Recognition and Understanding, 2003.
  ASRU'03. 2003 IEEE Workshop on}, pages 94--98. IEEE, 2003.

\bibitem[Luong and Manning(2016)]{luong2016achieving}
Minh-Thang Luong and Christopher~D Manning.
\newblock Achieving open vocabulary neural machine translation with hybrid
  word-character models.
\newblock \emph{arXiv preprint arXiv:1604.00788}, 2016.

\bibitem[Cartwright and Brent(1994)]{cartwright1994segmenting}
Timothy~Andrew Cartwright and Michael~R Brent.
\newblock Segmenting speech without a lexicon: The roles of phonotactics and
  speech source.
\newblock \emph{arXiv preprint cmp-lg/9412005}, 1994.

\bibitem[Goldwater et~al.(2006)Goldwater, Griffiths, and
  Johnson]{goldwater2006contextual}
Sharon Goldwater, Thomas~L Griffiths, and Mark Johnson.
\newblock Contextual dependencies in unsupervised word segmentation.
\newblock In \emph{Proceedings of the 21st International Conference on
  Computational Linguistics and the 44th annual meeting of the Association for
  Computational Linguistics}, pages 673--680. Association for Computational
  Linguistics, 2006.

\bibitem[Sak et~al.(2015)Sak, Senior, Rao, and Beaufays]{Sak2015FastAA}
Hasim Sak, Andrew~W. Senior, Kanishka Rao, and Françoise Beaufays.
\newblock Fast and accurate recurrent neural network acoustic models for speech
  recognition.
\newblock \emph{CoRR}, abs/1507.06947, 2015.

\bibitem[Kim et~al.(2016)Kim, Hori, and Watanabe]{kim2016joint}
Suyoun Kim, Takaaki Hori, and Shinji Watanabe.
\newblock Joint ctc-attention based end-to-end speech recognition using
  multi-task learning.
\newblock \emph{arXiv preprint arXiv:1609.06773}, 2016.

\bibitem[Kim and Rush(2016)]{kim2016sequence}
Yoon Kim and Alexander~M Rush.
\newblock Sequence-level knowledge distillation.
\newblock \emph{arXiv preprint arXiv:1606.07947}, 2016.

\bibitem[Miao et~al.(2015)Miao, Gowayyed, and Metze]{miao2015eesen}
Yajie Miao, Mohammad Gowayyed, and Florian Metze.
\newblock Eesen: End-to-end speech recognition using deep rnn models and
  wfst-based decoding.
\newblock In \emph{Automatic Speech Recognition and Understanding (ASRU), 2015
  IEEE Workshop on}, pages 167--174. IEEE, 2015.

\bibitem[Zhang et~al.(2016)Zhang, Chan, and Jaitly]{zhang2016very}
Yu~Zhang, William Chan, and Navdeep Jaitly.
\newblock Very deep convolutional networks for end-to-end speech recognition.
\newblock \emph{arXiv preprint arXiv:1610.03022}, 2016.

\bibitem[Zweig et~al.(2016{\natexlab{b}})Zweig, Yu, Droppo, and
  Stolcke]{geoffery2016ctc}
Geoffery Zweig, Ghengzhu Yu, Jasha Droppo, and Andreas Stolcke.
\newblock Advances in all-neural speech recognition.
\newblock \emph{arXiv preprint arXiv:1609.05935}, 2016{\natexlab{b}}.

\bibitem[Povey et~al.(2016)Povey, Peddinti, Galvez, Ghahrmani, Manohar, Na,
  Wang, and Khudanpur]{povey2016purely}
Daniel Povey, Vijayaditya Peddinti, Daniel Galvez, Pegah Ghahrmani, Vimal
  Manohar, Xingyu Na, Yiming Wang, and Sanjeev Khudanpur.
\newblock Purely sequence-trained neural networks for asr based on lattice-free
  mmi.
\newblock \emph{Submitted to Interspeech}, 2016.

\bibitem[Chorowski and Navdeep(2016)]{chorowski2016seq2seq}
Jan Chorowski and Jaitly Navdeep.
\newblock Towards better decoding and language model integration in sequence to
  sequence models.
\newblock \emph{arXiv preprint arXiv:1612.02695}, 2016.

\end{thebibliography}

\end{document}